\documentclass[12pt]{article}

\usepackage{sbc-template}
\usepackage{graphicx,url}
\usepackage[utf8]{inputenc}
\usepackage[brazil]{babel}

\usepackage[breaklinks]{hyperref}

\usepackage{float}
\usepackage{booktabs}
\usepackage{graphicx}
\usepackage{float}
\usepackage{array}
\usepackage{colortbl}
\usepackage{copyrightbox}
\usepackage{amsmath}
\usepackage{comment}
\usepackage{booktabs}

\title{Detecção da Psoríase Utilizando Visão Computacional: Uma Abordagem Comparativa Entre CNNs e Vision Transformers}

\author{Natanael Lucena\inst{1}, Fabio S. da Silva\inst{1}, Ricardo Rios\inst{1}}

\address{Escola Superior de Tecnologia \textemdash~ Universidade do Estado do Amazonas\\Manaus, AM, Brasil
  \email{\{nldm.eng20, fssilva, rrios\}@uea.edu.br}
}

\begin{document}

\maketitle

\begin{abstract}
This paper presents a comparison of the performance of Convolutional Neural Networks (CNNs) and \textit{Vision Transformers} (ViTs) in the task of multiclassifying images containing lesions of psoriasis and diseases similar to it. Models pre-trained on ImageNet were adapted to a specific data set. Both achieved high predictive metrics, but the ViTs stood out for their superior performance with smaller models. Dual Attention Vision Transformer-Base (DaViT-B) obtained the best results, with an f1-score of 96.4\%, and is recommended as the most efficient architecture for automated psoriasis detection. This article reinforces the potential of ViTs for medical image classification tasks.
\end{abstract}

\begin{resumo}
Esse artigo apresenta uma comparação de desempenho de Redes Neurais Convolucionais (CNNs) e \textit{Vision Transformers} (ViTs) na tarefa de multiclassificação de imagens contendo lesões de psoríase e de enfermidades similares a essa doença. Modelos pré-treinados no ImageNet foram adaptados a um conjunto de dados específico. Ambos alcançaram métricas preditivas elevadas, mas os ViTs se destacaram por apresentarem desempenho superior com modelos menores. O \textit{Dual Attention Vision Transformer-Base} (DaViT-B) obteve os melhores resultados, com um \textit{f1-score} de 96,4\%, e é recomendado como a arquitetura mais eficiente para detecção automatizada de psoríase. Esse artigo reforça o potencial dos ViTs para tarefas de classificação de imagens médicas.

\end{resumo}

\section{Introdução}

A psoríase é uma doença inflamatória da pele, crônica e sem cura, que afeta aproximadamente 3\% da população mundial, ou seja, cerca de 125 milhões de pessoas \cite{sbd_psoriase}. No Brasil, estima-se que 90\% da população desconhece a doença, e apenas 6\% reconhece suas lesões \cite{agenciabrasil2020}. O diagnóstico atual é realizado por dermatologistas com base em achados clínicos e pistas visuais \cite{dash2020}, sendo um processo subjetivo e demorado \cite{ji2022}, que pode exigir biópsias para confirmação \cite{rodrigues2009}. A complexidade do diagnóstico aumenta devido à semelhança visual da psoríase com outras doenças de pele, como dermatite, líquen plano e pitiríase rosada \cite{roslan2020}.

A aplicação de técnicas de visão computacional e inteligência artificial (IA) tem se mostrado promissora para auxiliar no diagnóstico de doenças de pele, incluindo a psoríase \cite{milani2023, olescki2021, silva2022}. Métodos baseados em aprendizado profundo (\textit{Deep Learning}), como Redes Neurais Convolucionais (CNNs) e \textit{Vision Transformers} (ViTs), têm alcançado desempenho próximo ou superior ao humano em tarefas de classificação de imagens \cite{wei2023skin, mohan2024enhancing, roslan2020, skindisease1}. Essas técnicas podem explorar características da pele que não são facilmente visíveis ao olho humano, oferecendo uma solução objetiva e potencialmente automatizada para o diagnóstico.

Esse artigo apresenta um estudo comparativo entre modelos de CNNs e ViTs para a tarefa de multi-classificação de imagens de pele com psoríase e doenças visualmente similares. O objetivo é identificar o modelo mais adequado para atuar como ferramenta de apoio ao diagnóstico dermatológico. Para isso, foram selecionados modelos de CNNs (EfficientNet-V2-L \cite{tan2021efficientnetv2}, ConvNeXt-L \cite{liu2022convnet} e Inception-v3 \cite{szegedy2016rethinking}) e ViTs (ViT-L/16 \cite{alexey2020image}, MaxViT-T \cite{tu2022maxvit} e DaViT-B \cite{ding2022davit}), pré-treinados no ImageNet \cite{deng2009imagenet} e adaptados a um conjunto de dados específico de imagens de psoríase e doenças similares.

Este trabalho expande a abordagem adotada em \cite{milani2023} de classificação binária para multi classificação, incluindo imagens de doenças similares, e a utilização de ViTs na comparação.

Esse artigo está organizado como segue. Na Seção \ref{Sec:TrabRelacionados}, são apresentados os trabalhos relacionados. A Seção \ref{Sec:MateriasMetodos} apresenta os materiais e métodos utilizados neste trabalho. Na Seção \ref{Sec:Resultados}, discutem-se os resultados obtidos a partir do treinamento dos modelos. Por fim, na Seção \ref{Sec:Conclusões} são apresentadas as conclusões do trabalho.

\section{Trabalhos Relacionados} \label{Sec:TrabRelacionados}

\cite{milani2023} focou na aplicação de CNNs para a detecção de psoríase. Foi comparado o desempenho de diferentes arquiteturas de CNN. Os autores utilizaram técnicas como validação cruzada K-Fold \cite{kfold2001} e aumento de dados. O modelo Inception-v3 destacou-se com um desempenho superior, alcançando uma acurácia de 97,5\%. Foi ressaltada a importância do aumento de dados para a generalização do modelo. Entretanto, o estudo foca exclusivamente na classificação binária das imagens de psoríase e pele saudável, o que simplifica o problema e favorece resultados otimistas. Além disso, compara três modelos clássicos de CNN — ResNet50 \cite{he2016deep}, Inception-v3 \cite{szegedy2016rethinking} e VGG19 \cite{simonyan2014very} — que, embora amplamente utilizados, são relativamente antigos.

\cite{zhao2020smart} utilizou CNNs para a identificação de psoríase a partir de um conjunto de dados estruturado de imagens clínicas, incluindo doenças de pele visualmente similares, como líquen plano e eczema.
O modelo Inception-v3 obteve o melhor desempenho, com uma AUC de $0,981 \pm 0,015$, superando outros modelos como Xception \cite{chollet2017xception} e DenseNet121 \cite{huang2017densely}. O sistema de IA demonstrou maior precisão e menor taxa de erros em comparação com dermatologistas humanos. No entanto, devido aos dados não estarem disponíveis publicamente, a comunidade científica pode enfrentar limitações para reproduzir o trabalho. Além disso, foi mencionado o uso de técnicas de aumento de dados, mas os autores não esclareceram se essas técnicas foram devidamente restritas ao conjunto de treinamento.

\cite{mohan2024enhancing} explorou o uso de modelos baseados em \textit{Transformers}, incluindo ViTs e \textit{Swin Transformers} \cite{liu2021swin}, para a classificação de doenças de pele, incluindo psoríase. Foi utilizado um conjunto de dados ampliado de 31 classes e alcançou uma precisão de teste de 96,48\% com o modelo DinoV2 \cite{oquab2023dinov2}, superando \textit{benchmarks} anteriores. A pesquisa concluiu que os \textit{Transformers} têm potencial para revolucionar o diagnóstico de doenças de pele. Os autores não explicam se o conjunto de dados foi ampliado ou refinado. Apesar de o artigo apontar que os modelos ViT do experimento obtiveram um maior desempenho em relação à CNN EfficientNetB2, não fica claro se o conjunto de dados utilizado contém imagens de doenças visualmente similares à psoríase. Neste trabalho, o conjunto de dados contempla imagens de psoríase e de doenças similares, o que aumenta o desafio da classificação.

\section{Materiais e Métodos} \label{Sec:MateriasMetodos}

Esse trabalho utiliza modelos baseados na técnica de Aprendizado Supervisionado para detectar a incidência de psoríase.

Para o treinamento dos modelos, fez-se o uso de um servidor composto de duas GPUs NVIDIA GeForce GTX 1080 Ti, cada uma com 11 GB de VRAM, um processador Intel(R) Core(TM) i7-8700 CPU @ 3,20 GHz, 56 GB de memória principal e 2,4 TB de memória secundária.

\subsection{Conjunto de Dados} \label{Subsec:ConjDados}

A obtenção de imagens médicas representou um desafio na construção da base de dados devido a questões éticas e à necessidade de preservar a privacidade dos pacientes. O acesso a imagens dermatológicas requer autorizações explícitas, o que dificulta a coleta e o compartilhamento. Por isso, o conjunto de imagens, de diferentes áreas do corpo, foi coletado a partir de repositórios de domínio público \cite{dermnetnz, dermis, dermatoweb, atlasdermatologico, danderm, hellenicdermatlas}, cujos dados de imagem podem ser utilizados para fins educativos e acadêmicos. Essas imagens possuem etiquetas de diagnóstico, validadas por dermatologistas certificados.

A coleta de imagens abrangeu lesões rotuladas como psoríase, dermatite, líquen plano e pitiríase rosada, resultando inicialmente em um total de 3.248 imagens. Imagens de áreas específicas, como unhas e couro cabeludo, foram desconsideradas por não apresentarem características visuais relevantes. Além disso, imagens desfocadas ou sem foco foram removidas, reduzindo o conjunto final para 2.781 imagens.

O conjunto de imagens de pele saudável, também de diversas áreas do corpo humano, foi coletado de dados NTU \cite{ntu2014} do \textit{Biometrics and Forensics Lab}. O total de imagens coletadas dessa fonte foi uma quantidade sucinta de 20 imagens. O restante das imagens de pele saudável foi obtido do Google Images®, resultando em um total de 1.176 imagens de pele saudável.

Devido à coleta de dados de múltiplas fontes, um processo de refinamento foi necessário para eliminar imagens duplicadas. Um \textit{script} em Python, utilizando o algoritmo \emph{dHash}, foi implementado para detectar e remover duplicatas. O procedimento identificou e excluiu redundâncias, garantindo um conjunto de dados único e mais representativo.

\begin{figure}
    \centering
    \includegraphics[width=0.9\linewidth]{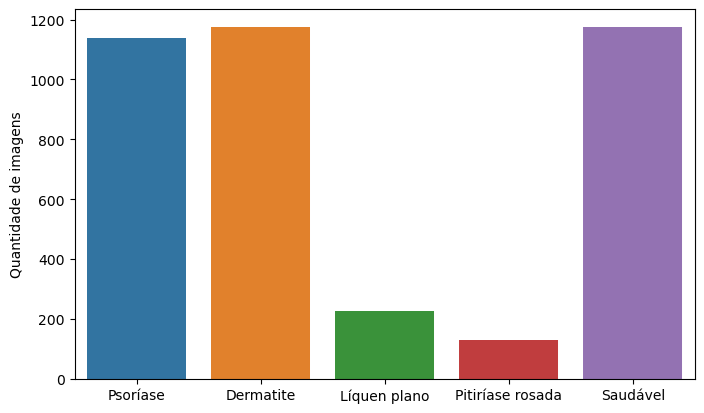}
    \caption{Distribuição do conjunto de dados por classe.}
    \label{fig:datasetDistribution}
\end{figure}

Após todas as etapas de refinamento, o conjunto final de dados totalizou 3.843 imagens distribuídas entre as classes de psoríase, dermatite, líquen plano, pitiríase rosada e pele saudável, conforme apresenta o gráfico da Figura \ref{fig:datasetDistribution}.

\subsection{Particionamento do Conjunto de Dados} \label{Subsec:PartConjDados}

O particionamento foi realizado em duas etapas. Na primeira, o conjunto de dados foi dividido de modo que 80\% dos dados foram destinados a participar da validação cruzada, representando os dados de treino e validação. A outra parte, 20\%, foi separada exclusivamente para compor os dados de teste, que não participam da validação cruzada e nem da etapa de treino. Nessa etapa foi empregada a técnica de Amostragem Estratificada \cite{meng2013}, que garante que os subgrupos estejam adequadamente representados no conjunto de treinamento.

Na segunda etapa, foi utilizada a validação cruzada K-\textit{Fold} \cite{kfold2001} no subconjunto de treino e validação. Optou-se por utilizar cinco (05) \textit{folds}, sendo quatro para treinamento e um exclusivo para validação. Durante cada iteração do processo de validação, um dos \textit{folds} de treinamento é selecionado para validação, permitindo que o modelo seja avaliado em diferentes subconjuntos de dados. Esse método contribui para minimizar o risco de \textit{overfitting} em apenas um conjunto específico e aumenta a confiabilidade e a precisão dos resultados obtidos \cite{Geron2019}.

\subsection{Pré-Processamento das Imagens} \label{Subsec:PreProcImagens}

Todas as imagens tiveram suas intensidades de pixel normalizadas, de modo a colocá-las em uma escala entre 0 (zero) e 1 (um). Além disso, o conjunto de dados destinado ao treinamento passou por um processo de aumento artificial de dados. Essa técnica incluiu rotações incrementais das imagens em ângulos aleatórios entre 0$^{\circ}$ e 20$^{\circ}$, além da aplicação de giros tanto horizontais quanto verticais. Por fim, as imagens foram redimensionadas para o tamanho especificado por cada modelo pré-treinado no ImageNet.

\subsection{Modelos de CNN e ViT} \label{Subsec:ModCNNs}

Optou-se por utilizar a versão pré-treinada (a partir da base do ImageNet) de cada modelo, disponível nas bibliotecas TorchVision \cite{marcel2010torchvision} e Timm \cite{rw2019timm}.


A tabela \ref{tab:eachModel} apresenta o tamanho e a quantidade de parâmetros de cada modelo, além de suas respectivas acurácias \textit{top-1} e \textit{top-5} na base de dados do ImageNet.

\begin{table}[ht]
\centering
\begin{tabular}{|c|c|c|c|c|c|}
\hline
\textbf{Modelo} & \textbf{Categoria} & \begin{tabular}[x]{@{}c@{}}\textbf{Tamanho}\\ \textbf{(MB)}\end{tabular} & \textbf{Parâmetros} & \begin{tabular}[x]{@{}c@{}}\textbf{Acurácia}\\ \textbf{\textit{top-1} (\%)}\end{tabular} & \begin{tabular}[x]{@{}c@{}}\textbf{Acurácia}\\ \textbf{\textit{top-5} (\%)}\end{tabular}  \\ \hline
Inception-v3     & CNN & 104   & 27,2M  & 77,3\%  & 93,5\%  \\ \hline
EfficientNetV2-L & CNN & 455   & 118,5M & 85,8\%  & 97,8\%  \\ \hline
ConvNeXt-L       & CNN & 755   & 197,8M & 84,4\%  & 97,0\%  \\ \hline
ViT-L/16         & ViT & 1.161 & 304,3M & 79,7\%  & 94,6\%  \\ \hline
MaxViT-T         & ViT & 118   & 30,9M  & 83,7\%  & 96,7\%  \\ \hline
DaViT-B          & ViT & 352   & 88,0M  & 84,6\%  & 96,9\%  \\ \hline
\end{tabular}
\caption{Modelos selecionados e suas respectivas acurácias no conjunto de dados ImageNet \cite{marcel2010torchvision, rw2019timm}.}
\label{tab:eachModel}
\end{table}

Os modelos foram selecionados com base em suas características específicas, buscando avaliar os resultados obtidos em diferentes configurações e levando em conta seu desempenho na competição ImageNet. Um fator comum entre eles é a ampla disponibilidade em bibliotecas de aprendizado profundo, como TorchVision e Timm, o que facilita sua implementação.

\subsection{Transferência de Aprendizado}\label{Cap04:transferenciaAprendizadoMetodologia}

Treinar uma rede neural profunda completa para uma tarefa específica demanda elevados recursos computacionais e um grande volume de dados. Por isso, é comum recorrer a modelos pré-treinados, que podem ser utilizados como ponto de partida ou como extratores de características para o problema em questão.

A estratégia definida para transferência de aprendizado foi a de utilizar um modelo pré-treinado como extrator de características, devido às suas vantagens significativas de eficiência, simplicidade e redução do risco de \textit{overfitting}.

Essa abordagem consiste em empregar um modelo pré-treinado, geralmente desenvolvido com base em um conjunto de dados extenso e genérico, como, por exemplo, o ImageNet, para obter as características das novas imagens. Posteriormente, essas características extraídas são utilizadas para treinar um modelo adaptado a um conjunto de dados menor e mais específico. Nesse processo, a última camada do modelo é substituída por uma camada densa com o número de unidades correspondente ao total de classes do novo conjunto de dados \cite{milani2023}.

\subsection{Configurações do Treinamento} \label{SubSec:ParamTreinamentos}

Durante o treinamento dos modelos, foi definido um critério para acompanhar as métricas de perda (\textit{loss}) e acurácia (\textit{accuracy}). Esse critério de otimização foi definido para o caso de o modelo apresentar melhorias na taxa de aprendizado (\textit{learning rate}). A seguir, serão apresentadas as configurações utilizadas em todos os treinamentos realizados.

\begin{itemize}
    \item Otimizador: utilizou-se o otimizador \textit{AdaMax}, que é uma versão do Adam \cite{adam2014} mais robusta para grandes gradientes e valores de parâmetros elevados.

    \item \textit{ReduceLROnPlateau}: técnica de redução da taxa de aprendizado mediante platô. A métrica utilizada para identificar o platô da taxa de aprendizagem é a perda (\textit{loss}) do conjunto de validação. A redução aplicada tem um fator de 1e\textsuperscript{-3}, e a paciência (\textit{patience}) foi definida como 3, ou seja, o modelo espera por três épocas sem melhorias antes de diminuir a taxa de aprendizagem usando o fator.

    \item \textit{Early Stopping}: a acurácia do conjunto de validação é a métrica utilizada para avaliar o desempenho do modelo, e a paciência foi definida como 7 épocas.

    \item \textit{Checkpoint}: utilizou-se as funções \textit{save} e \textit{load\_state\_dict} para salvar e carregar, respectivamente, os parâmetros do modelo.

    \item \textit{Batch size}: utilizou-se um \textit{batch} de tamanho 32 levando em consideração os recursos computacionais disponíveis e o tempo de treinamento.

    \item Épocas: Para esse experimento, definiu-se 50 épocas.

\end{itemize}

\subsection{Métricas de Avaliação} \label{Subsec:Metricas}

A matriz de confusão é representada por uma matriz quadrada, utilizada para comparar as previsões do modelo com os valores reais (verdadeiros). Os valores localizados na diagonal principal correspondem às previsões corretas, enquanto as demais posições representam os erros do modelo. Ela apresenta quatro categorias de rótulos:

\begin{itemize}
    \item Verdadeiro Negativo (TN): quantidade de amostras corretamente classificadas como pertencentes à classe negativa.

    \item Verdadeiro Positivo (TP): quantidade de amostras corretamente classificadas como pertencentes à classe positiva.

    \item Falso Negativo (FN): número de amostras da classe positiva incorretamente classificadas como negativas.

    \item Falso Positivo (FP): número de amostras da classe negativa incorretamente classificadas como positivas.
\end{itemize}

Com isso, a partir dos parâmetros que compõem a matriz de confusão, é possível obter cada métrica de avaliação. Neste trabalho foram utilizadas métricas ponderadas.

As métricas ponderadas calculam as métricas individuais para cada classe, mas ponderam essas métricas pelo número de amostras de cada classe antes de calcular a média final. Desse modo, as métricas ponderadas conseguem equilibrar a análise de classes minoritárias e majoritárias, sendo úteis para conjuntos de dados desbalanceados onde é desejável refletir o impacto real das classes no desempenho geral \cite{luque2019impact}.

Nas fórmulas de cada métrica a seguir, a variável $N$ representa o número total de classes:

\begin{itemize}
    \item Acurácia: mede a proporção de previsões corretas em relação ao total de amostras avaliadas, conforme a equação \ref{eqn:acuracia}.

    \begin{equation} \label{eqn:acuracia}
        \text{Acurácia} = \frac{1}{N} \sum_{i=1}^{N} \frac{TP_i + TN_i}{TP_i + TN_i + FP_i + FN_i}
    \end{equation}

    \item Precisão: é a métrica utilizada para representar quantos casos de verdadeiros positivos são identificados corretamente pelo modelo, conforme a equação \ref{eqn:precisao}.

    \begin{equation} \label{eqn:precisao}
        \text{Precisão} = \frac{1}{N} \sum_{i=1}^{N} \frac{TP_i}{TP_i + FP_i}
    \end{equation}

    \item Revocação (\textit{recall}): também chamada de \emph{sensibilidade}, é utilizada para representar as amostras positivas que foram corretamente rotuladas, conforme a equação \ref{eqn:revocacao}.

    \begin{equation} \label{eqn:revocacao}
        \text{Revocação} = \frac{1}{N} \sum_{i=1}^{N} \frac{TP_i}{TP_i + FN_i}
    \end{equation}

    \item \textit{F1-score}: é calculado a partir da média harmônica da precisão e sensibilidade, conforme a equação \ref{eqn:f1Score}. O \textit{f1-score} é especialmente útil em contextos onde há um desbalanceamento entre as classes, pois penaliza modelos que apresentam alta precisão, mas baixa revocação, ou vice-versa.

    \begin{equation} \label{eqn:f1Score}
        \textit{F1-score} = \frac{1}{N} \sum_{i=1}^{N} \frac{2 \times \text{Precisão} \times \text{Revocação}}{\text{Precisão} + \text{Revocação}}
    \end{equation}

\end{itemize}

\section{Resultados e Discussão} \label{Sec:Resultados}

A Tabela \ref{tab:modelResults} apresenta as métricas de cada um dos seis modelos selecionados. Cada métrica foi calculada a partir da média ponderada considerando os resultados da validação cruzada K-\textit{fold}.

\newcommand{\specialcell}[2][c]{%
  \begin{tabular}[#1]{@{}c@{}}#2\end{tabular}}

\begin{table}[ht]
\centering
\caption{Métricas gerais em termos de média ponderada.}
\bigskip
\begin{tabular}{*6c}\toprule
 \textbf{Modelo} & \textbf{Parâmetros} & \textbf{Acurácia} & \textbf{Precisão}  & \textbf{Revocação} & \textbf{F1-\textit{score}}\\\midrule
 Inception-v3     & 27,2M  & $94,5\%$ & $94,6\%$ & $94,5\%$ & $94,5\%$ \\
 EfficientNetV2-L & 118,5M & $95,4\%$ & $95,5\%$ & $95,4\%$ & $\textbf{95,4\%}$ \\
 ConvNeXt-L       & 197,8M & $95,6\%$ & $95,6\%$ & $95,6\%$ & $\textbf{\textcolor{blue}{95,5}}\%$ \\ \midrule
 ViT-L/16         & 304,3M & $88,5\%$ & $88,0\%$ & $88,5\%$ & $87,8\%$ \\
 MaxViT-T         & 30,9M  & $\textbf{96,0\%}$ & $\textbf{96,1\%}$ & $\textbf{96,0\%}$ & $\textbf{96,0\%}$ \\
 DaViT-B          & 88,0M  & $\textbf{\textcolor{red}{96,4\%}}$ & $\textbf{\textcolor{red}{96,4\%}}$ & $\textbf{\textcolor{red}{96,4\%}}$ & $\textbf{\textcolor{red}{96,4\%}}$ \\ \bottomrule
\end{tabular}\\
\label{tab:modelResults}
\end{table}

Dentre os modelos de CNN, ConvNeXt-L e EfficientNetV2-L alcançaram um \textit{f1-score} de \emph{95,4\%} e 95,5\%, respectivamente. No entanto, esse desempenho foi obtido a custo de uma quantidade de parâmetros significativamente elevada de ambos os modelos (118,5 milhões e 197,8 milhões, respectivamente). Uma vez que a quantidade de parâmetros de uma rede neural pode influenciar diretamente no seu custo computacional \cite{lecun1989optimal}, essa quantidade elevada de parâmetros pode indicar uma maior limitação quanto à sua implementação prática em ambientes com recursos limitados. Por outro lado, o modelo Inception-v3 se apresenta como uma alternativa mais econômica em termos de recursos computacionais, dado que conta com apenas 27,2 milhões de parâmetros, ao custo de um desempenho de predição ligeiramente inferior, mas ainda notável, com um \textit{f1-score} de 94,5\%.

Dentre os modelos de ViT, MaxViT-T e DaViT-B alcançaram as métricas mais altas (acurácia, precisão, revocação e \textit{f1-score}) ao custo de quantidades de parâmetros relativamente baixas (30,9 milhões e 88,0 milhões, respectivamente). O modelo ViT-L/16, apesar de sua grande contagem de parâmetros (304,3 milhões), obteve o pior desempenho em comparação com os outros modelos, alcançando um \textit{f1-score} inferior, de 87,8\%. Esse desempenho pode ser justificado pelo fato de o modelo não ter sido bem ajustado para o conjunto de dados do ImageNet, conforme aponta \cite{tan2021efficientnetv2}.

É possível observar o desempenho superior dos modelos DaViT-B e MaxViT-T para a tarefa proposta, principalmente por ambos terem alcançado métricas de predição excepcionais, mesmo sendo modelos relativamente leves. Entretanto, apesar de ser ligeiramente mais pesado que o MaxViT-T, o DaViT-B acertou 97\% das classificações de psoríase, conforme apontado em sua matriz de confusão. Esse fator, somado a sua conquista de maior \textit{f1-score} (96,4\%) dentre os modelos, consolida o DaViT-B como o modelo com maior potencial para atuar como ferramenta de auxílio no diagnóstico da psoríase. 

\begin{figure}[ht]
    \includegraphics[width=0.982\linewidth]{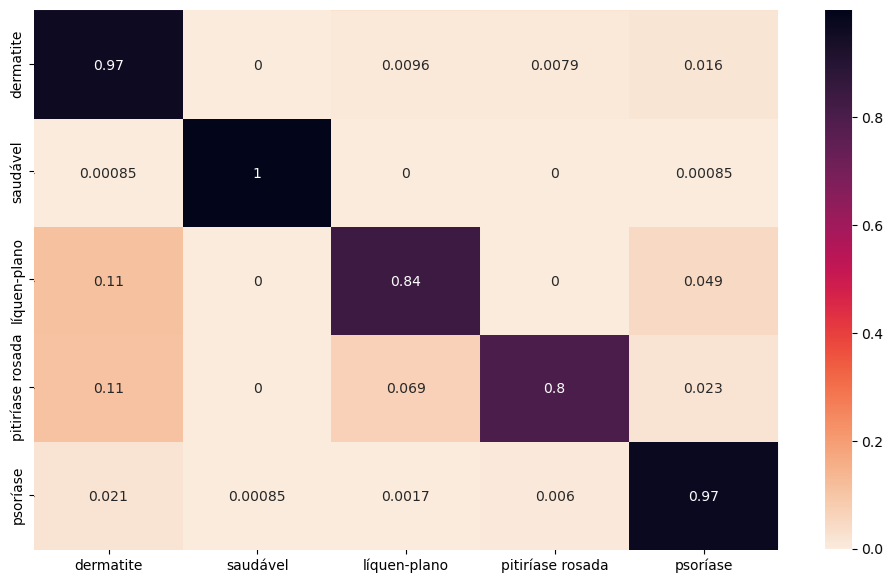}
    \caption{Matriz de Confusão do modelo DaViT-B.}
    \label{fig:davitConfusionMatrix}
\end{figure}

Os dados da matriz de confusão do Davit-B (Figura \ref{fig:davitConfusionMatrix}) indicam que a principal dificuldade de classificação para o modelo residiu na semelhança visual dessas condições com a psoríase e, especialmente, com a dermatite.
O modelo classificou corretamente 84\% das imagens de líquen plano, sendo os principais erros a confusão com dermatite (11\%) e psoríase (4,9\%); outras doenças, como a pitiríase rosada (6,9\% dos seus casos), também foram erroneamente classificadas como líquen plano. Para a pitiríase rosada, a taxa de acerto foi de 80\%, com confusões primárias com dermatite (11\%) e líquen plano (6,9\%).

Finalmente, evidencia-se que a dermatite constitui o principal fator de confusão para ambas as patologias em questão. Ademais, uma dificuldade notável foi observada na diferenciação mútua entre líquen plano e pitiríase rosada. Esses erros podem ser justificados pelo nível notável de similaridade visual das lesões cutâneas, que impõe limitações ao desempenho diagnóstico do DaViT-B.

\section{Conclusões} \label{Sec:Conclusões}

O experimento aponta que ambas as categorias de modelos (CNN e ViT) apresentam um grande potencial para a resolução da tarefa proposta, com todos os modelos, com exceção do ViT-L/16, alcançando um f1-\textit{score} superior a 94\%. Em particular, foi observado que DaViT-B melhor eficiência na detecção da psoríase, sendo o modelo selecionado como o mais indicado para a tarefa.

A implementação de modelos como o DaViT-B em ambientes clínicos para auxiliar dermatologistas na detecção da psoríase é promissora, uma vez que o desempenho robusto do modelo em diferenciar psoríase de outras condições de pele visualmente similares demonstra seu potencial para reduzir diagnósticos equivocados. Isso favorece a detecção mais refinada da doença, para que tratamentos mais precisos e eficazes sejam prescritos.

A alta similaridade visual das lesões cutâneas pode justificar a principal confusão na classificação de líquen plano e pitiríase rosada pelo modelo DaViT-B, que ocorre com a dermatite e entre si. Entretanto, é possível que haja mais fatores que corroborem com essa confusão e, por isso, reconhece-se a necessidade de investigações futuras mais aprofundadas para elucidar os fatores que contribuem para a essas confusões e para o desenvolvimento de estratégias mitigatórias, como o enriquecimento do conjunto de dados.

Para futuros trabalhos planeja-se expandir o conjunto de dados com imagens de um leque maior de doenças de pele que se assemelhem à psoríase (o melanoma e outros tipos de câncer de pele); expandir a análise dos erros de predição apresentados por todos os modelos para as classes Líquen Plano, Pitiríase Rosada e Dermatite; explorar a transferência de aprendizado do ajuste fino do modelo pré-treinado (Seção \ref{Cap04:transferenciaAprendizadoMetodologia}); investigar a utilização do otimizador \textit{Nesterov-accelerated Adaptative Moment Estimation} (Nadam) \cite{dozat2016incorporating}. Além disso, pretende-se definir e coletar métricas de avaliação do custo computacional de cada modelo (métrica de predições por segundo ou do tempo de uma predição), verificar o desempenho, com ajuste fino, de modelos multimodais populares e modernos (Llama 3.2 \cite{dubey2024llama}, Gemini 1.5 \cite{team2024gemini} e GPT-4o \cite{hurst2024gpt}), e desenvolver um sistema que utilize o DaViT-B para o auxílio ao diagnóstico da psoríase e coletar métricas de testes desse sistema em ambientes de uso prático.

\section{Agradecimentos}

Os autores agradecem pelo apoio do Laboratório de Sistemas Inteligentes (LSI) da Universidade do Estado do Amazonas (UEA), na pessoa da Professora do Núcleo de Computação (NuComp) da Escola Superior de Tecnologia (EST). Dra. Elloá B. Guedes.


\medskip

\begin{thebibliography}{}

    \bibitem[Alexey 2020]{alexey2020image}
    Alexey, D. (2020).
    \newblock An image is worth 16x16 words: Transformers for image recognition at
    scale.
    \newblock {\em arXiv preprint arXiv: 2010.11929}.

    \bibitem[Atlas Dermatologico 2024]{atlasdermatologico}
    Atlas Dermatologico (2024).
    \newblock Dermatology atlas.
    \newblock \url{https://atlasdermatologico.com.br/}.
    \newblock Acesso em: 19/04/2024.

    \bibitem[Bin~Ji 2022]{ji2022}
    Bin~Ji, Yiyi~Wang, D.~Z. (2022).
    \newblock Automatic detection and evaluation of nail psoriasis based on deep
    learning: a preliminary application and exploration.
    \newblock {\em SPIE International Conference on Computer Application and
        Information Security}.

    \bibitem[Brasil 2020]{agenciabrasil2020}
    Brasil, A. (2020).
    \newblock Estudo mostra que mais de 90\% da população desconhecem a
    psoríase.
    \newblock
    \url{https://agenciabrasil.ebc.com.br/saude/noticia/2020-11/estudo-mostra-que-mais-de-90-da-populacao-desconhecem-psoriase}.
    \newblock Acesso em: 29/04/2024.

    \bibitem[Chollet 2017]{chollet2017xception}
    Chollet, F. (2017).
    \newblock Xception: Deep learning with depthwise separable convolutions.
    \newblock In {\em Proceedings of the IEEE conference on computer vision and
        pattern recognition}, pages 1251--1258.

    \bibitem[Danderm 2017]{danderm}
    Danderm (2017).
    \newblock Atlas of dermatology.
    \newblock \url{https://www.danderm.dk/atlas/}.
    \newblock Acesso em: 17/04/2024.

    \bibitem[Dash et~al. 2020]{dash2020}
    Dash, M. et~al. (2020).
    \newblock A cascaded deep convolution neural network based cadx system for
    psoriasis lesion segmentation and severity assessment.
    \newblock {\em Applied Soft Computing}.

    \bibitem[de~Dermatologia 2023]{sbd_psoriase}
    de~Dermatologia, S.~B. (2023).
    \newblock O que é a psoríase?
    \newblock \url{https://www.sbd.org.br/psoriase/}.
    \newblock Acesso em: 29/04/2024.

    \bibitem[Deng et~al. 2009]{deng2009imagenet}
    Deng, J., Dong, W., Socher, R., Li, L.-J., Li, K., and Fei-Fei, L. (2009).
    \newblock Imagenet: A large-scale hierarchical image database.
    \newblock In {\em 2009 IEEE conference on computer vision and pattern
        recognition}, pages 248--255. Ieee.

    \bibitem[Dermatoweb 2002]{dermatoweb}
    Dermatoweb (2002).
    \newblock Web docente de dermatologia.
    \newblock \url{http://dermatoweb.udl.es/}.
    \newblock Acesso em: 22/04/2024.

    \bibitem[DermIS 2022]{dermis}
    DermIS (2022).
    \newblock Dermatology information system.
    \newblock \url{https://www.dermis.net/}.
    \newblock Acesso em: 26/04/2024.

    \bibitem[DermNetNZ 2024]{dermnetnz}
    DermNetNZ (2024).
    \newblock The worlds leading free dermatology website.
    \newblock \url{https://dermnetnz.org/}.
    \newblock Acesso em: 29/04/2024.

    \bibitem[Ding et~al. 2022]{ding2022davit}
    Ding, M., Xiao, B., Codella, N., Luo, P., Wang, J., and Yuan, L. (2022).
    \newblock Davit: Dual attention vision transformers.
    \newblock In {\em European conference on computer vision}, pages 74--92.
    Springer.

    \bibitem[Dozat 2016]{dozat2016incorporating}
    Dozat, T. (2016).
    \newblock Incorporating nesterov momentum into adam.
    \newblock In {\em Proceedings of the 4th International Conference on Learning
        Representations, Workshop Track}, pages 1--4.

    \bibitem[Dubey et~al. 2024]{dubey2024llama}
    Dubey, A., Jauhri, A., Pandey, A., Kadian, A., Al-Dahle, A., Letman, A.,
    Mathur, A., Schelten, A., Yang, A., Fan, A., et~al. (2024).
    \newblock The llama 3 herd of models.
    \newblock {\em arXiv preprint arXiv:2407.21783}.

    \bibitem[G{\'e}ron 2019]{Geron2019}
    G{\'e}ron, A. (2019).
    \newblock {\em M{\~a}os {\`A} Obra: Aprendizado de M{\'a}quina com
        {Scikit-Learn} {E} {TensorFlow}}.
    \newblock Alta Books.

    \bibitem[He et~al. 2016]{he2016deep}
    He, K., Zhang, X., Ren, S., and Sun, J. (2016).
    \newblock Deep residual learning for image recognition.
    \newblock In {\em Proceedings of the IEEE conference on computer vision and
        pattern recognition}, pages 770--778.

    \bibitem[Hellenic Dermatological Atlas 2011]{hellenicdermatlas}
    Hellenic Dermatological Atlas (2011).
    \newblock For health professionals and public.
    \newblock \url{http://www.hellenicdermatlas.com/en/}.
    \newblock Acesso em: 16/04/2024.

    \bibitem[Huang et~al. 2017]{huang2017densely}
    Huang, G., Liu, Z., Van Der~Maaten, L., and Weinberger, K.~Q. (2017).
    \newblock Densely connected convolutional networks.
    \newblock In {\em Proceedings of the IEEE conference on computer vision and
        pattern recognition}, pages 4700--4708.

    \bibitem[Hurst et~al. 2024]{hurst2024gpt}
    Hurst, A., Lerer, A., Goucher, A.~P., Perelman, A., Ramesh, A., Clark, A.,
    Ostrow, A., Welihinda, A., Hayes, A., Radford, A., et~al. (2024).
    \newblock Gpt-4o system card.
    \newblock {\em arXiv preprint arXiv:2410.21276}.

    \bibitem[Huynh et~al. 2014]{ntu2014}
    Huynh, N.~Q., Xu, X., Kong, A. W.~K., and Subbiah, S. (2014).
    \newblock A preliminary report on a full-body imaging system for effectively
    collecting and processing biometric traits of prisoners.
    \newblock In {\em 2014 IEEE Symposium on Computational Intelligence in
        Biometrics and Identity Management (CIBIM)}.

    \bibitem[Kingma and Ba 2014]{adam2014}
    Kingma, D. and Ba, J. (2014).
    \newblock Adam: A method for stochastic optimization.
    \newblock {\em International Conference on Learning Representations}.

    \bibitem[Kohavi 1995]{kfold2001}
    Kohavi, R. (1995).
    \newblock A study of cross-validation and bootstrap for accuracy estimation and
    model selection.
    \newblock In {\em Proceedings of the 14th International Joint Conference on
        Artificial Intelligence - Volume 2}. Morgan Kaufmann Publishers Inc.

    \bibitem[LeCun et~al. 1989]{lecun1989optimal}
    LeCun, Y., Denker, J., and Solla, S. (1989).
    \newblock Optimal brain damage.
    \newblock {\em Advances in neural information processing systems}, 2.

    \bibitem[Liu et~al. 2021]{liu2021swin}
    Liu, Z., Lin, Y., Cao, Y., Hu, H., Wei, Y., Zhang, Z., Lin, S., and Guo, B.
    (2021).
    \newblock Swin transformer: Hierarchical vision transformer using shifted
    windows.
    \newblock In {\em Proceedings of the IEEE/CVF international conference on
        computer vision}, pages 10012--10022.

    \bibitem[Liu et~al. 2022]{liu2022convnet}
    Liu, Z., Mao, H., Wu, C.-Y., Feichtenhofer, C., Darrell, T., and Xie, S.
    (2022).
    \newblock A convnet for the 2020s.
    \newblock In {\em Proceedings of the IEEE/CVF conference on computer vision and
        pattern recognition}, pages 11976--11986.

    \bibitem[Luque et~al. 2019]{luque2019impact}
    Luque, A., Carrasco, A., Mart{\'\i}n, A., and de~Las~Heras, A. (2019).
    \newblock The impact of class imbalance in classification performance metrics
    based on the binary confusion matrix.
    \newblock {\em Pattern Recognition}, 91:216--231.

    \bibitem[Maduranga and Nandasena 2022]{skindisease1}
    Maduranga, P. and Nandasena, D. (2022).
    \newblock Mobile-based skin disease diagnosis system using convolutional neural
    networks (cnn).
    \newblock {\em International Journal of Image, Graphics and Signal Processing},
    14:47--57.

    \bibitem[Marcel and Rodriguez 2010]{marcel2010torchvision}
    Marcel, S. and Rodriguez, Y. (2010).
    \newblock Torchvision the machine-vision package of torch.
    \newblock \url{https://pytorch.org/vision/}.

    \bibitem[Meng 2013]{meng2013}
    Meng, X. (2013).
    \newblock Scalable simple random sampling and stratified sampling.
    \newblock In Dasgupta, S. and McAllester, D., editors, {\em Proceedings of the
        30th International Conference on Machine Learning}, volume~28 of {\em
        Proceedings of Machine Learning Research}, pages 531--539, Atlanta, Georgia,
    USA. PMLR.

    \bibitem[Milani et~al. 2023]{milani2023}
    Milani, A. et~al. (2023).
    \newblock A deep learning application for psoriasis detection.
    \newblock {\em Anais do Encontro Nacional de Inteligência Artificial e
        Computacional (ENIAC)}, pages 315--329.

    \bibitem[Mohan et~al. 2024]{mohan2024enhancing}
    Mohan, J., Sivasubramanian, A., Sowmya, V., and Vinayakumar, R. (2024).
    \newblock Enhancing skin disease classification leveraging transformer-based
    deep learning architectures and explainable ai.
    \newblock {\em arXiv preprint arXiv:2407.14757}.

    \bibitem[Olescki 2021]{olescki2021}
    Olescki, G. (2021).
    \newblock Detecção de tromboembolia pulmonar utilizando redes neurais
    convolucionais e extração de características.
    \newblock {\em Anais do XXI Simpósio Brasileiro de Computação Aplicada à
        Saúde}, pages 381--391.

    \bibitem[Oquab et~al. 2023]{oquab2023dinov2}
    Oquab, M., Darcet, T., Moutakanni, T., Vo, H., Szafraniec, M., Khalidov, V.,
    Fernandez, P., Haziza, D., Massa, F., El-Nouby, A., et~al. (2023).
    \newblock Dinov2: Learning robust visual features without supervision.
    \newblock {\em arXiv preprint arXiv:2304.07193}.

    \bibitem[Rodrigues and Teixeira 2009]{rodrigues2009}
    Rodrigues, A.~P. and Teixeira, R.~M. (2009).
    \newblock Desvendando a psoríase.
    \newblock {\em RBAC}.

    \bibitem[Roslan et~al. 2020]{roslan2020}
    Roslan, R., Razly, I., Sabri, B., and Ibrahim, Z. (2020).
    \newblock Evaluation of psoriasis skin disease classification using
    convolutional neural network.
    \newblock {\em IAES International Journal of Artificial Intelligence (IJ-AI)},
    9:349.

    \bibitem[Silva et~al. 2022]{silva2022}
    Silva, G. et~al. (2022).
    \newblock Cardiac arrhythmia detection in ecg signals using graph convolutional
    network.
    \newblock {\em Anais do XXII Simpósio Brasileiro de Computação Aplicada à
        Saúde}, pages 25--35.

    \bibitem[Simonyan 2014]{simonyan2014very}
    Simonyan, K. (2014).
    \newblock Very deep convolutional networks for large-scale image recognition.
    \newblock {\em arXiv preprint arXiv:1409.1556}.

    \bibitem[Szegedy et~al. 2016]{szegedy2016rethinking}
    Szegedy, C., Vanhoucke, V., Ioffe, S., Shlens, J., and Wojna, Z. (2016).
    \newblock Rethinking the inception architecture for computer vision.
    \newblock In {\em Proceedings of the IEEE conference on computer vision and
        pattern recognition}, pages 2818--2826.

    \bibitem[Tan and Le 2021]{tan2021efficientnetv2}
    Tan, M. and Le, Q. (2021).
    \newblock Efficientnetv2: Smaller models and faster training.
    \newblock In {\em International conference on machine learning}, pages
    10096--10106. PMLR.

    \bibitem[Team et~al. 2024]{team2024gemini}
    Team, G., Georgiev, P., Lei, V.~I., Burnell, R., Bai, L., Gulati, A., Tanzer,
    G., Vincent, D., Pan, Z., Wang, S., et~al. (2024).
    \newblock Gemini 1.5: Unlocking multimodal understanding across millions of
    tokens of context.
    \newblock {\em arXiv preprint arXiv:2403.05530}.

    \bibitem[Tu et~al. 2022]{tu2022maxvit}
    Tu, Z., Talebi, H., Zhang, H., Yang, F., Milanfar, P., Bovik, A., and Li, Y.
    (2022).
    \newblock Maxvit: Multi-axis vision transformer.
    \newblock In {\em European conference on computer vision}, pages 459--479.
    Springer.

    \bibitem[Wei et~al. 2023]{wei2023skin}
    Wei, M., Wu, Q., Ji, H., Wang, J., Lyu, T., Liu, J., and Zhao, L. (2023).
    \newblock A skin disease classification model based on densenet and convnext
    fusion.
    \newblock {\em Electronics}, 12(2):438.

    \bibitem[Wightman 2019]{rw2019timm}
    Wightman, R. (2019).
    \newblock Pytorch image models.
    \newblock \url{https://github.com/rwightman/pytorch-image-models}.

    \bibitem[Zhao et~al. 2020]{zhao2020smart}
    Zhao, S., Xie, B., Li, Y., Zhao, X.-y., Kuang, Y., Su, J., He, X.-y., Wu, X.,
    Fan, W., Huang, K., et~al. (2020).
    \newblock Smart identification of psoriasis by images using convolutional
    neural networks: a case study in china.
    \newblock {\em Journal of the European Academy of Dermatology and Venereology},
    34(3):518--524.

\end{thebibliography}
\end{document}